# Beyond the Cloud:
# Assessing the Benefits and Drawbacks of Local LLM Deployment for Translators

Peter Sandrini

*Department of Translation Studies, University of Innsbruck*

---

**Abstract:**

*The rapid proliferation of Large Language Models (LLMs) presents both opportunities and challenges for the translation field. While commercial, cloud-based AI chatbots have garnered significant attention in translation studies, concerns regarding data privacy, security, and equitable access necessitate exploration of alternative deployment models. This paper investigates the feasibility and performance of locally deployable, free LLMs as a viable alternative to proprietary, cloud-based AI solutions. This study evaluates three open-source LLMs – Llama 3, Gemma 2, Mixtral 8x7B – installed on CPU-based platforms and compared against commercially available online chatbots. The evaluation focuses on functional performance rather than a comparative analysis of human-machine translation quality, an area already subject to extensive research. The platforms assessed were GPT4All, Llamafile, and Ollama, chosen for their accessibility and ease of use across various operating systems. While local deployment introduces its own challenges, the benefits of enhanced data control, improved privacy, and reduced dependency on cloud services are compelling. The findings of this study contribute to a growing body of knowledge concerning the democratization of AI technology and inform future research and development efforts aimed at making LLMs more accessible and practical for a wider range of users, specifically focusing on the needs of individual translators and small businesses.*

---

# 1 Introduction

Large Language Models (LLMs) have emerged as a transformative technology with the potential to revolutionize numerous fields, including translation. These sophisticated AI systems, capable of processing and generating human-quality text, offer significant advantages in terms of speed, accuracy, and scalability. In the field of translation studies most publications focus on the use and integration of commercial online AI chatbots such as ChatGPT (e.g., Preciado et al, 2025; Rivas Ginel/Moorkens, 2024; Chao/Kim, 2023; Miller/Thompson, 2024).

However, the current landscape of LLM development and deployment is dominated by a few powerful tech companies, raising concerns about data privacy, security, and the equitable distribution of AI benefits. As the concentration of power in the hands of a few corporations intensifies, the risk of a future where AI is controlled by a select few becomes increasingly apparent. This scenario could lead to a lack of transparency, accountability, and diversity in AI development, potentially exacerbating existing societal inequalities. Even if those prominent AI online services are offered for free by the big tech companies, they are used to gathering user data, in line with the well-known saying: "When something is free, you are the product" (Richard Serra 1973).

To counter this trend, the development of free and Open Source LLMs is essential: "The development of Open Source generative AI is crucial for promoting transparency, accountability, and inclusion in the creation and deployment of these powerful technologies" (Merilehto, 2024: 1). By fostering a more decentralized and collaborative approach to AI research and development, we can ensure that the benefits of this technology are widely accessible. Translators can use such LLMs without fear of violating their NDA with clients. Moreover, they are not dependent on software companies or internet connections. Educational institutions do not need to enter into expensive licensing agreements or beg for educational licenses. In addition, new applications can be developed on the basis of free models.

To address these challenges, the Free Software Foundation (FSF) has advocated for the adoption of free and Open Source principles in AI development and is working on a definition of free and open AI which integrates the four essential freedoms: the freedom to run, study, share, and improve software. However, such a definition for Open Source AI may prove difficult: "defining 'Open Source' for foundation models (FMs) has proven tricky" (Basdevant et al., 2024: 1). Moreover, Open Source in AI may also involve some new risks, like allowing malicious actors to disable safeguards against misuse, or introducing new dangerous capabilities via fine-tuning (Seger et al., 2023: 2). The initiatives *Open Weights* (Open Weights Definition) and *Explainable AI* (XAI) (Alvarez-Melis and Jaakola, 2017) may constitute a first step in the direction of completely Open Source AI (OSI, 2025).

While we agree with Open Source principles, it is practically impossible to understand and reproduce the inner working of artificial neural nets, as they are aptly named black boxes (Federal Office for Information Security, 2024: 9). So, in the following remarks we concentrate on freely available LLMs without adhering to the strict requirements of the four freedoms outlined by the FSF. We chose available Ai platforms that can be installed on a local computer and which do not use subscription schemes, credits (as do e.g., ChatGPT, FreedomGPT) or other pay options.

By investigating the limitations and potential of CPU-based AI systems in contrast to well-known online systems, we seek to inform future research and development efforts aimed at making AI technology more accessible and practical for wider range of users. The ultimate goal of this study is to evaluate the feasibility of local LLMs and to test their performance for translation against much more powerful commercial online Chatbots. What this study does not aim at, is a comprehensive quality evaluation or a human-machine comparison, considering that this is the focus of many other studies (e.g., Jozić, 2024; Calvo-Ferrer, 2023; Briva-Iglesias et al., 2024; Giampieri, 2024; Choi and Kim, 2023).

# 2 Available free AI-Containers

A platform where such free LLMs can be downloaded is Hugging Face: A Startup and ecosystem dedicated to democratizing AI with an Open Source Transformers library. However, the use of such LLMs requires specific knowledge in programming languages such as Python, and requires the use of Python libraries such as LangChain, PyTorch, LlamaIndex, as well as competence in computer infrastructure and software engineering. A few projects try to simplify this process and provide a container which allows the integration of one or more LLMs, giving users the opportunity to choose which LLM they want to use for a specific use case. Thus, translators are able to profit from AI-tools without being AI engineers or programmers. For this study three freely available platforms or frameworks were selected.

## 2.1 GPT4ALL

GPT4All by Nomic AI represents a platform which allows you to run a great variety of local models on CPUs and GPUs. It can also access locally stored texts and documents with sensitive information. This platform supports the longest list of freely available LLMs. For this study we used version GPT4All 3.7 until the end of January 2025, then version 3.8.

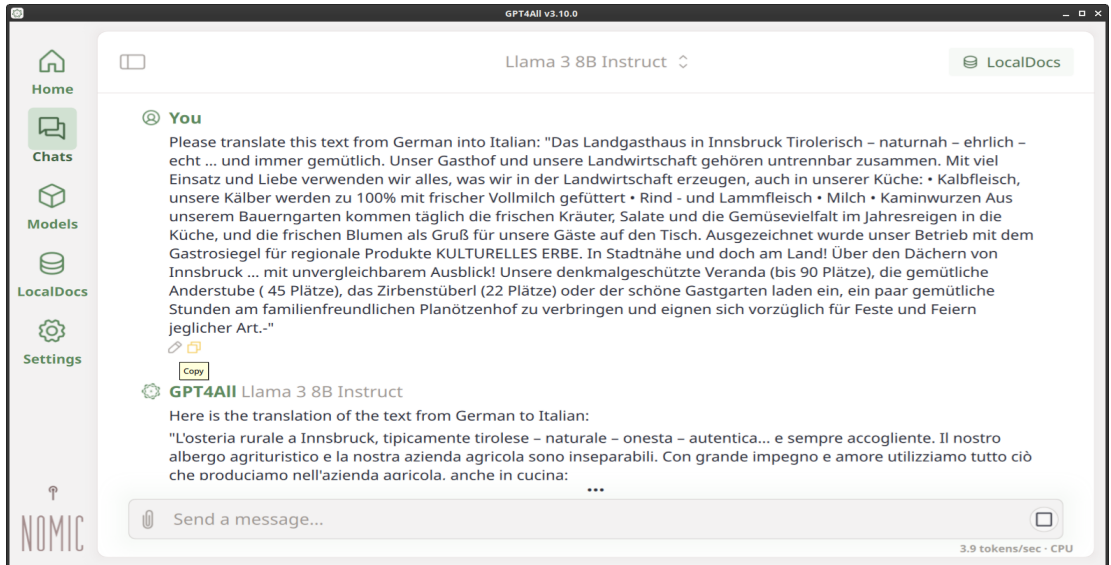

*ChatGPT user interface*

## 2.2 Llamafile

Mozilla Innovation Project Llamafile is an Open Source initiative that collapses all the complexity of a full-stack LLM chatbot down to a single file that runs on six operating systems. As a single file, it makes distributing and running models painless by keeping its process simple and straightforward. With minimal complexity Llamafile also supports a list of different LLMs. Llamafile models were all downloaded in December 2024.

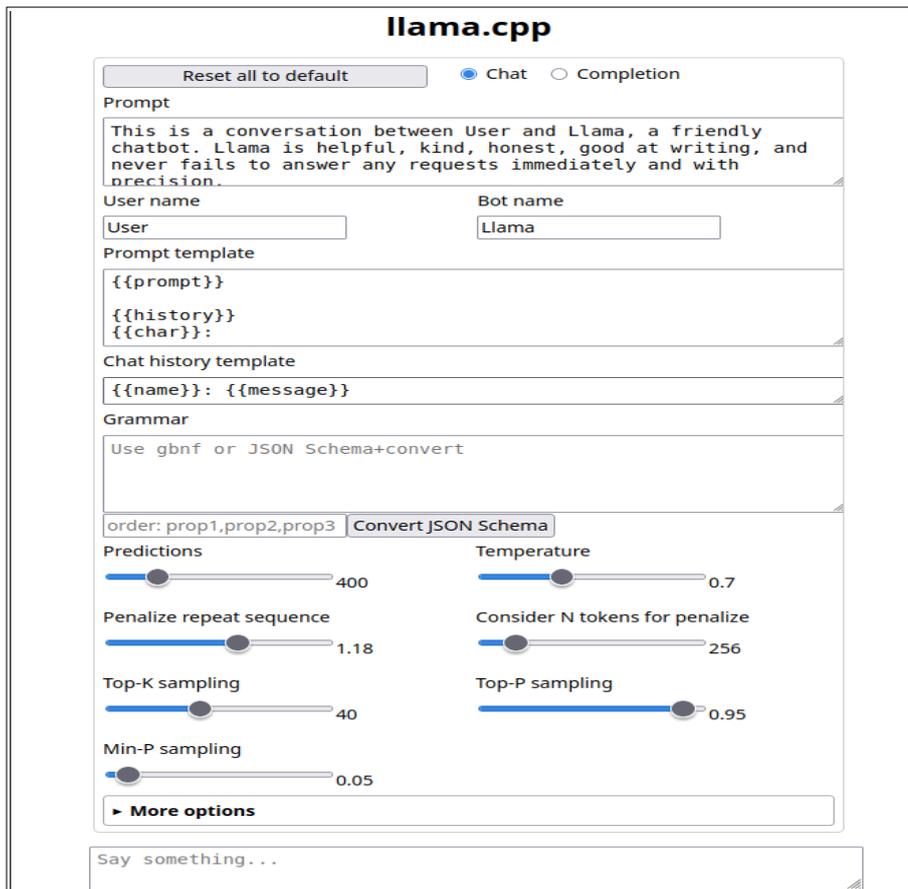

*Llamafile user interface*

## 2.3 Ollama

Ollama is another platform which allows users to run, customize, and interact with LLMs directly on their own computers locally with all the advantages of improved privacy and reduced dependency on cloud providers. Ollama provides access to multiple LLMs on their GitHub website. The following test was done with Ollama 0.5.7 and the browser interface PageAssist.

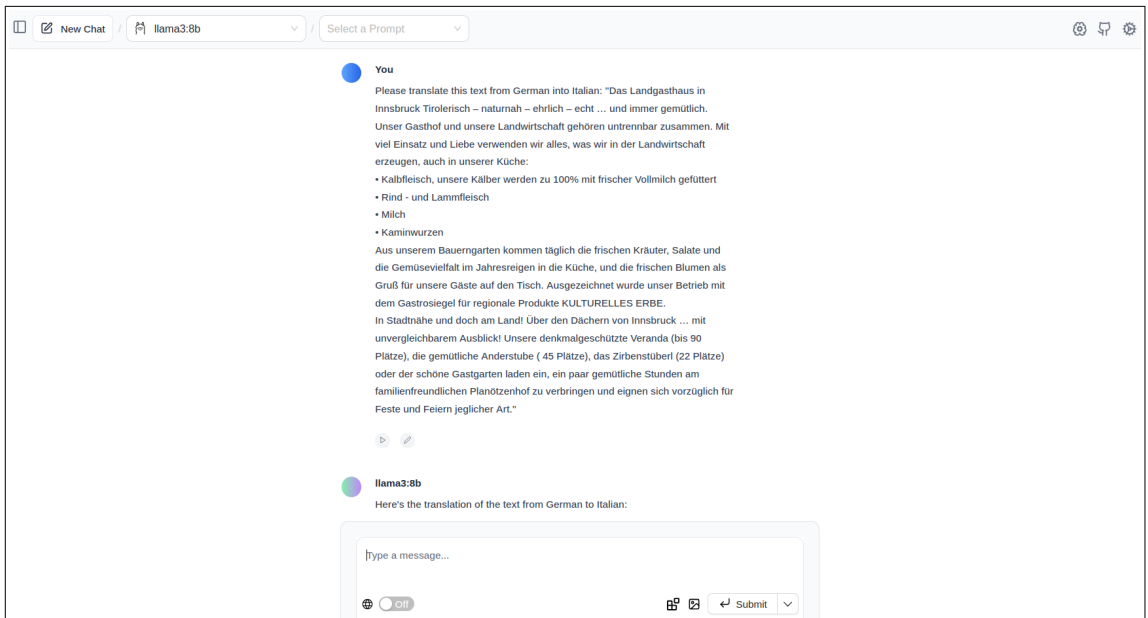

*Ollama with PageAssist user interface*

In addition to these readily available platforms there are a few other frameworks for the use of local LLMs which were not used for the present study.

# 2.4 Other systems

**PrivateGPT**

One of the first platforms to be developed. In principle Open Source, free to use for personal use, but complicated to install due to dependencies and various packages.

**AnythingLLM**

AnythingLLM introduces another open-source platform for AI models. It is a desktop application primarily designed to interface with online chatbots, with the added capability of integrating locally hosted large language models (LLMs).

**LM-Studio**

In principle Open Source, free to use for personal use, but rather difficult to install due to dependencies and additional packages.

**Cheshire Cat**

The Cheshire Cat is an Open Source, hackable and production-ready framework that allows developing AI agents on top of LLMs.

**LocalAI**

LocalAI represents another Open Source platform able to run LLMs locally, it does not require a GPU.

**Opus-Cat-MT**

A dedicated NMT translation engine

**DLTranslator and OmegaT**

A test version for the integration of LLMs into the translation memory system OmegaT which works as a simple MT plugin.

# 3 Local LLMs

By hosting models on personal hardware, users retain control over their data, mitigating potential privacy risks associated with cloud-based solutions. However, local deployment presents its own set of challenges. It requires significant upfront investment in hardware, as well as ongoing maintenance to ensure system reliability and performance. For individual translators and small businesses, the technical expertise and financial resources required for local deployment can be substantial barriers.

The decisive factor in selecting LLMs for this research was their accessibility across all platforms under consideration. Each platform provides a curated collection of models, optimized for its specific environment. While the exact number of models varies across platforms, the availability of the same LLMs was essential for our comparative analysis. This accessibility allowed us to conduct a quality comparison of different platform architectures and their performance across various tasks.

Therefore, we performed our tests on three platforms with three models per platform. The models available on all three chosen platforms were:

- Llama 3, a pre-trained and instruction-fine-tuned language model with 8B parameters by Meta AI. While Meta also has a 70B parameter version, we did not use it because latency times were too long.
- Gemma 2, a lightweight, state-of-the-art open model by Google AI, built using the same research and technology as the Gemini models. Gemma 2 also offers a variety of parameter sizes, but for the same reasons, we chose the 27B version.
- Mixtral 8x7B, an open model developed by Mistral AI, a French company, which boasts fast inference times and high performance.

Although these LLMs are accessible across all platforms, there are subtle variations in their size and versioning. These differences arise from the need to optimize the models for the specific requirements and constraints of each platform. Many other LLMs are used on these platforms, but they are not available in all environments. Some models are geared toward coding or mathematics, others toward vision and graphics, while a few boast multilingual capabilities, such as the AYA or the Mixtral 8x7B models.

# 4 Test setup

We conducted our tests using the Italian-German language pair due to our familiarity and expertise with these languages. A deep understanding of these languages allows for a more accurate and insightful evaluation of translation results. Additionally, by avoiding English, we aimed to obtain a more realistic assessment of the models' capabilities with less-represented languages. While Italian and German may be underrepresented in training data compared to English, we believe this language pair is sufficiently well-represented to yield satisfactory results, unlike significantly less-resourced language combinations such as Hungarian-Danish or Czech-Polish.[1]

Contrary to some suggestions (e.g., Garcia, 2009 "Beyond Translation Memory"), we maintain that professional translators still utilize Translation Environment Tools (TEnTs) equipped with integrated Translation Memory, Terminology Management, Machine Translation, Quality Assessment, and other functionalities (see ELIS 2024). To evaluate the potential of Large Language Models (LLMs), our initial tests involved preliminary preparation tasks, such as translating entire texts to create a translation memory and conducting upstream terminology work. The final tests focused on sending individual segments from a typical TEnT workflow to the LLM for translation.

This approach has major implications for latency. During the preliminary phase, latency is less critical, as the entire text is processed at once, along with terminology extraction. Translators can perform other tasks in the meantime. However, LLM integration within the translation memory workflow is extremely time-critical. Short latency times are crucial to maintain a smooth TM workflow and avoid disrupting the translator's productivity. Translators often work in a flow state, and any delay in receiving the LLM-generated translation proposal can break this flow and reduce efficiency. We tested this in the segment-level translation test. These two approaches may be regarded as alternatives, as LLM-translated sentences can be used within a CAT tool, just as pre-translated TMX files can serve "as a source of inspiration (53%)," which seems to be the most reported use of ChatGPT according to a study (Rivas Ginel/Moorkens, 2024: 268).

---

1     Meta for example states that "over 5% of the Llama 3 pretraining dataset consists of high-quality non-English data that covers over 30 languages" (https://ai.meta.com/blog/meta-llama-3). Google used "Primarily English-language content" (https://ai.google.dev/gemma/docs/core/model_card_2) for the training of Gemma 2. Only MistralAI declares multilingual support for its 8x7B model: "Mixtral 8x7B masters French, German, Spanish, Italian, and English" (https://mistral.ai/news/mixtral-of-experts).

One major difficulty in conducting such tests is the extreme volatility of applications and language models due to the fast-paced development in a "tumultuous environment" (Monzó-Nebot/Taza-Fuster, 2024: 8). Our analysis involved six tests:

1. Translation of a short text with a simple prompt
2. Translation of the same short text with a more detailed prompt
3. Conversion of the output into a TMX translation memory
4. Terminology extraction using a small specialized corpus
5. Sentence translation for use in a TEnT with a simple prompt
6. Sentence translation for use in a TEnT with a more detailed prompt

Each experiment was performed on the three platforms with the three different LLMs, for a total of 54 tests conducted during the first two months of 2025. All measured outcomes refer to a standard desktop PC system with an Intel i7-12700 CPU, an integrated Intel graphics chip, and 32 GB of RAM.

## 4.1 Text samples

The text samples used in this research come from two distinct domains. We chose a field where creative writing is in greater demand, namely tourism marketing (for the complete short text translation), and a field where precise terminology and accuracy are more important, namely corporate law (for terminology extraction and sentence-level translation). The marketing text was taken from the website of a local Austrian restaurant:

> Das Landgasthaus in Innsbruck
> Tirolerisch – naturnah – ehrlich – echt … und immer gemütlich
> Unser Gasthof und unsere Landwirtschaft gehören untrennbar zusammen. Mit viel Einsatz und Liebe verwenden wir alles, was wir in der Landwirtschaft erzeugen, auch in unserer Küche:
> • Kalbfleisch, unsere Kälber werden zu 100% mit frischer Vollmilch gefüttert
> • Rind - und Lammfleisch
> • Milch
> • Kaminwurzen
> Aus unserem Bauerngarten kommen täglich die frischen Kräuter, Salate und die Gemüsevielfalt im Jahresreigen in die Küche, und die frischen Blumen als Gruß für unsere Gäste auf den Tisch. Ausgezeichnet wurde unser Betrieb mit dem Gastrosiegel für regionale Produkte KULTURELLES ERBE.
> In Stadtnähe und doch am Land! Über den Dächern von Innsbruck … mit unvergleichbarem Ausblick! Unsere denkmalgeschützte Veranda (bis 90 Plätze), die gemütliche Anderstube ( 45 Plätze), das Zirbenstüberl (22 Plätze) oder der schöne Gastgarten laden ein, ein paar gemütliche Stunden am familienfreundlichen Planötzenhof zu verbringen und eignen sich vorzüglich für Feste und Feiern jeglicher Art.

For the terminology extraction task, we prepared a small, highly specialized parallel corpus. It comprised ten original Italian and ten Austrian examples of bylaws for limited liability companies. The Austrian bylaws totaled 331 KB and 41,406 words, while the Italian bylaws comprised 317 KB and 47,098 words. This corpus was unaligned. Using this small corpus, we evaluated the effectiveness of integrating external knowledge. This integration can be achieved through various methods, such as uploading texts, embedding them directly into prompts, or utilizing the retrieval-augmented generation (RAG) capabilities of the platforms.

For the sentence-level translation, we used excerpts from the bylaws of limited liability companies from our corpus and requested translations from Italian into German:

> Le partecipazioni dei soci, con il consenso di tutti i soci, possono essere determinate anche in misura non proporzionale ai rispettivi conferimenti, sia in sede di costituzione che di modifiche del capitale sociale.

## 4.2 Preparation Phase

Due to the CPU-based setup with local models, latency was a major concern for our tests. Thus, we divided the experiment into two phases based on the importance of the time factor. The first phase encompasses all necessary and helpful processes for the actual translation that can be performed beforehand, where time constraints are not crucial: the preparation phase (tests 1-4). The last two tests concern the translation process itself, where latency is critical (tests 5 and 6).

The first test involved translating a short marketing text for the restaurant Planötzenhof in the Alps. We investigated how well the system translates the German text into a viable Italian marketing webpage and how long this takes.

Test 1. Simple prompt (5.2): "Translate this from German into Italian," as with machine translation.

Test 2. Elaborate prompt (5.3): Specification of text type, function, and prospective use.

Test 3: Transforming the result into a translation memory in TMX format (5.4): Does the segmentation make sense? Is the format correct? Can the TMX file be reused in a translation environment (TEnT)? How long does it take?

Test 4: Terminology extraction (5.5). We provided the large language model (LLM) with a series of texts in the source and target languages and requested a bilingual term list. How easy is it to provide texts to the system? How many terms are extracted? How long does it take?

## 4.3 TM-Workflow

To find out whether local models would be useful when integrated into the typical workflow within a translation environment tool. If the framework allowed embedding of external knowledge, we used this functionality; in this case, we compared the output with and without integrated text samples. We used a segment from an Italian legal text: How well does the system perform in translating the Italian sentence into a viable German legal text? What is the latency?

Test 5. Simple prompt (5.6.1): "Translate this from Italian into German" as with MT.

Test 6. Elaborate prompt (5.6.2): Specification of text type, function, source and target legal systems, and prospective use.

# 5 Test Results

## 5.1 Quality Evaluation

Translation quality is a multifaceted and relative concept, significantly influenced by various contextual factors. This complexity is reflected in the definition provided by Koby et al.: "A quality translation demonstrates accuracy and fluency required for the audience and purpose and complies with all other specifications negotiated between the requester and provider, taking into account end-user needs" (Koby et al., 2014: 416). Furthermore, human evaluation can exhibit low inter-annotator agreement, indicating a degree of subjectivity inherent in such assessments.

This study does not aim for an absolute evaluation of the quality of local large language models (LLMs). Instead, its primary objective is to investigate the practical applicability of these LLMs for practitioners and freelance translators, specifically examining their performance in relation to online offerings from major technology companies. It is crucial to acknowledge that even commercially available artificial intelligence (AI) models are not without limitations. As Giampieri (2024: 16) observes, "Several scholars have analyzed the reliability of chatbot-based translations and have mostly found that their outputs are affected by limitations and drawbacks". Consequently, this research does not evaluate these commercial AI chatbots directly.

For the translation tasks, the online platform MATEO (Machine Translation Evaluation Online: Vanroy et al., 2023), developed at the University of Ghent, was utilized. Translations generated by the two best known online systems, ChatGPT and Gemini, served as the reference standard against which the output of the local models was compared. The automatic evaluation process employed six distinct metrics: BERTScore, BLEURT, COMET, BLEU, ChrF, and TER. These metrics are similarity measures and assess quality on a scale ranging from 0 to 100, where 0 represents the lowest quality and 100 signifies the highest. TER (Translation Edit Rate) is a distance measure and indicates the number of editing operations required to align the machine translation with the reference translation. Therefore, a higher TER score corresponds to lower quality, while a lower TER score signifies higher quality. However, it is important to note that "quality" in this context refers to relative similarity to the output of the reference system.

We acknowledge that automatic results often do not perfectly correlate with human judgment: "For this reason, it's best to be wary of drawing conclusions about MT quality based solely on automatic evaluation" (Moorkens et al., 2025: 84). Therefore, the output of the local models under evaluation was not assessed against an ideal or perfect translation. Instead, their performance is compared to that of widely used online models to determine their relative utility for practical applications. This approach is supported by the observation that "One difficulty in using these automatic measures is that their output is not meaningful except to compare one system against another" (Snover et al., 2006: 224).

For the evaluation of translation memory (TMX) generation task, however, the study relies on a formal assessment judging segmentation, alignment, and file format as well as adherence to the TMX standard. For terminology extraction we use a human-based comparative assessment conducted by the author of the study.

## 5.2 Simple translation task

Initially, we performed a simple translation of the marketing text using the following simple prompt:

**Please translate this text from German into Italian**

The output from ChatGPT (version 4.0 mini) and Gemini (version 1.5 Flash) served as the reference standard against which the output from our three local models was compared. The MATEO platform was used for automatic evaluation.

### 5.2.1 Llamafile

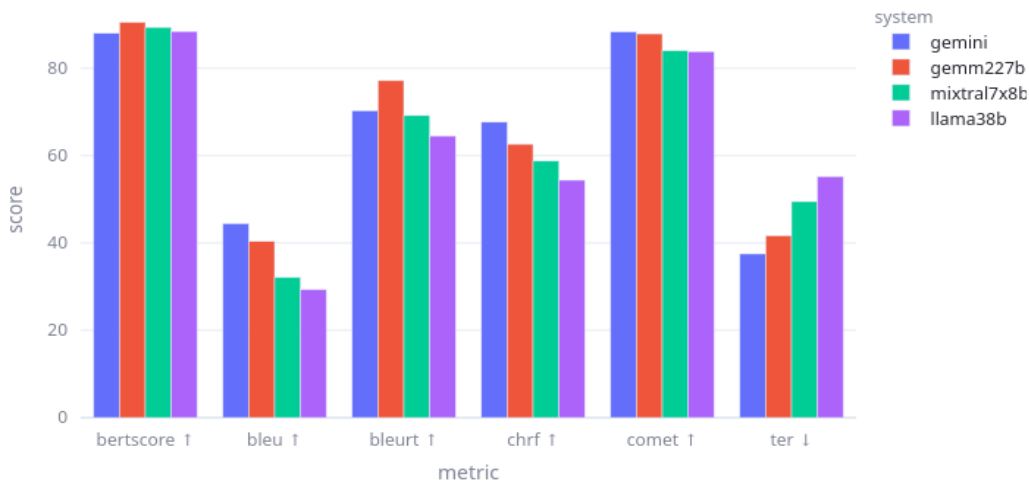

*Results with ChatGPT as reference*

| ChatGPT 4o mini | | | | | | | |
|---|---|---|---|---|---|---|---|
| **model** | **min.** | **bertscore** | **bleurt** | **comet** | **BLEU** | **chrF2** | **TER** |
| Gemini | 7,2 sec | 88,01 | 70,31 | 88,34 | 44,44 | 67,67 | 37,5 |
| Gemma2 27b | 5,13 | 90,53 | 77,18 | 87,89 | 40,34 | 62,64 | 41,67 |
| Mixtral 8x7b | 3,39 | 89,39 | 69,18 | 84,04 | 32,11 | 58,77 | 49,48 |
| Lama3 8b | 1,48 | 88,46 | 64,51 | 83,82 | 29,32 | 54,39 | 55,21 |

The results indicate that the smallest model, Llama 3 (8B), produced the lowest quality output, as evidenced by its lowest BLEU score and highest TER score. Conversely, Gemini and the largest model, Gemma 2 (27B), performed very similarly. This similarity in performance between Gemma 2 and Gemini may be attributed to the fact that Gemma is developed by Google and likely trained on comparable data, resulting in output that closely resembles that of Gemini.

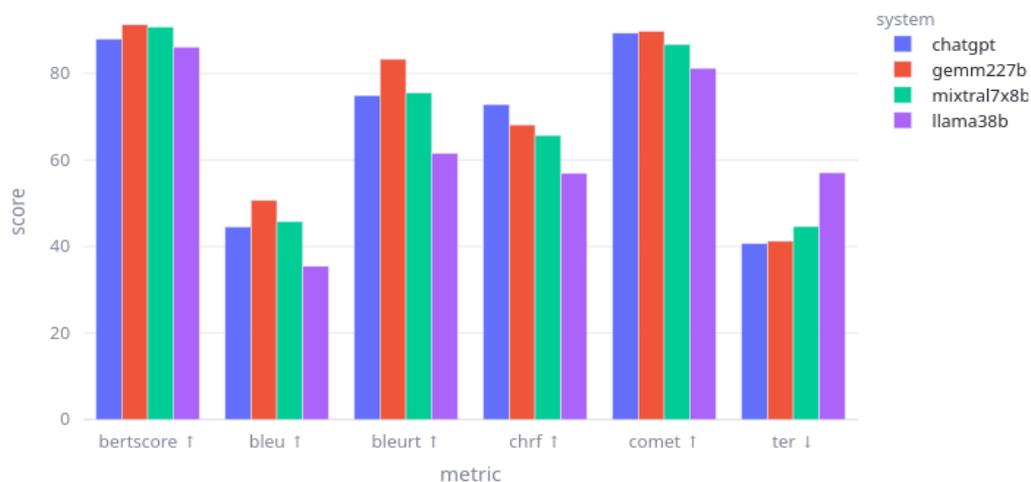

*Results with Gemini as reference*

| Gemini 1.5 Flash | | | | | | | |
|---|---|---|---|---|---|---|---|
| **model** | **min.** | **bertscore** | **bleurt** | **comet** | **BLEU** | **chrF2** | **TER** |
| ChatGPT | 7,4 sec | 88,01 | 74,94 | 89,36 | 44,57 | 72,85 | 40,68 |
| Gemma2 27b | 5,13 | 91,33 | 83,36 | 89,76 | 50,72 | 68,16 | 41,24 |
| Mixtral 8x7b | 3,39 | 90,79 | 75,57 | 86,77 | 45,79 | 65,7 | 44,63 |
| Lama3 8b | 1,48 | 86,13 | 61,57 | 81,22 | 35,43 | 56,99 | 57,06 |

Using Gemini as the reference yields similar results: the smallest model, Llama 3 (8B), consistently scores lowest across all metrics, while Gemma 2 (27B) either outperforms ChatGPT or achieves nearly equivalent scores. This again likely stems from the close relationship between Gemma 2 and the Gemini reference translation.

However, significant differences in latency were observed across both tests. While the online chatbots exhibited rapid response times of approximately 7 seconds, the latency of our local models was directly correlated with the number of parameters and model size. Larger models exhibited slower response times: Gemma 2, with 27 billion parameters and a size of 22.5 GB, had the longest latency. Mixtral (8x7 billion parameters, 30.03 GB), on the other hand, offered a favorable balance between latency, size, and relative translation quality.

### *5.2.2 Ollama*

We repeated the test for the Ollama platform with the same text and prompts.

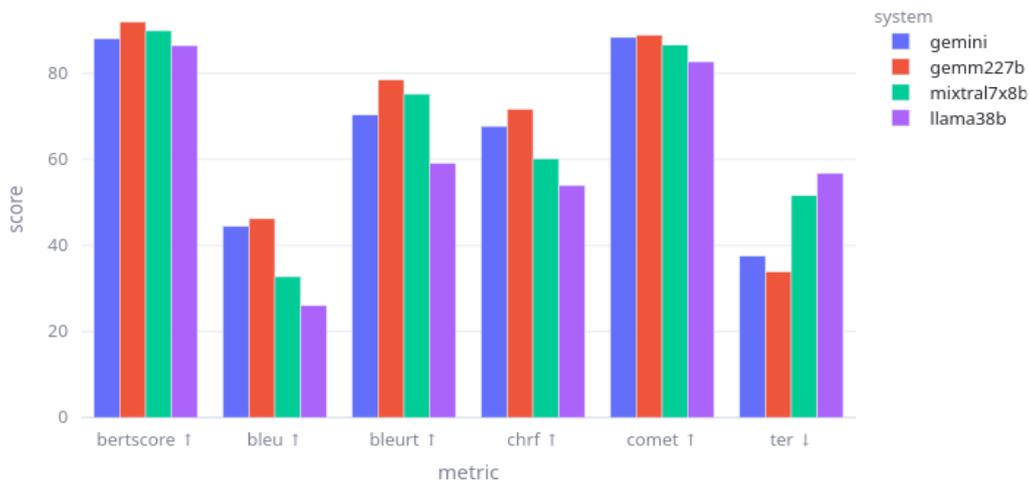

*Results with ChatGPT as reference*

| ChatGPT 4o mini | | | | | | | |
| model | min. | bertscore | bleurt | comet | BLEU | chrF2 | TER |
| --- | --- | --- | --- | --- | --- | --- | --- |
| Gemini | 7,2 sec | 88,01 | 70,31 | 88,34 | 44,44 | 67,67 | 37,5 |
| Gemma2 27b | 4,16 | 91,92 | 78,46 | 88,86 | 46,21 | 71,6 | 33,85 |
| Mixtral 8x7b | 3,2 | 89,88 | 75,17 | 86,53 | 32,74 | 60,14 | 51,56 |
| Lama3 8b | 1,36 | 86,43 | 59,04 | 82,67 | 26 | 53,95 | 56,77 |

Once again, the results reveal a significant difference in latency between online and local models. Gemma 2 (27B) outperformed Gemini across all metrics, indicating that Gemma 2's translation was more similar to ChatGPT's than to Gemini's. Llama 3 (8B) produced the lowest quality output.

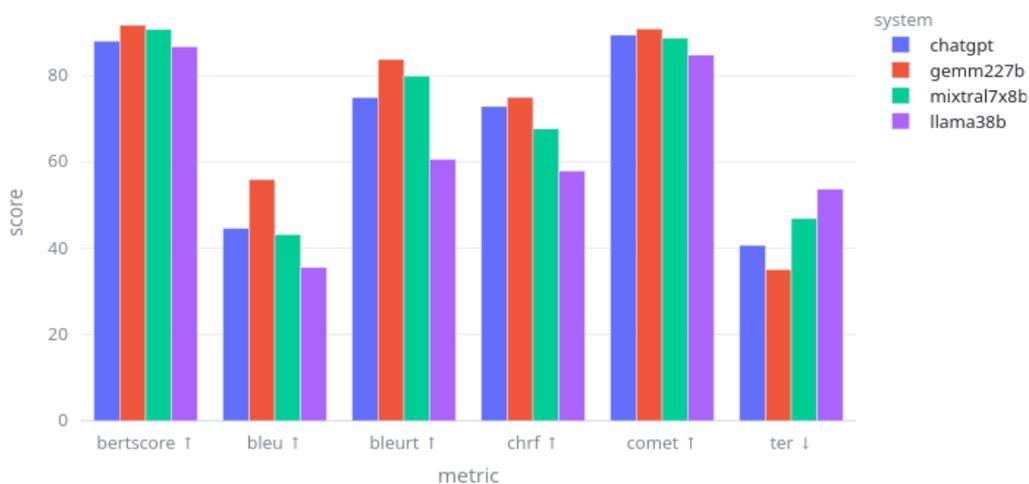

*Results with Gemini as reference*

| Gemini 1.5. Flash | | | | | | | |
| model | min. | bertscore | bleurt | comet | BLEU | chrF2 | TER |
| --- | --- | --- | --- | --- | --- | --- | --- |
| ChatGPT | 7,5 sec | 88,01 | 74,94 | 89,36 | 44,57 | 72,85 | 40,68 |
| Gemma2 27b | 4,16 | 91,68 | 83,73 | 90,84 | 55,92 | 74,97 | 35,03 |
| Mixtral 8x7b | 3,2 | 90,66 | 79,91 | 88,69 | 43,15 | 67,63 | 46,89 |
| Lama3 8b | 1,36 | 86,66 | 60,6 | 84,73 | 35,52 | 57,88 | 53,67 |

When the reference was changed to Gemini, the results indicated that, unsurprisingly, Gemma 2 (27B) was more similar to Gemini than to ChatGPT. The smallest model again scored lowest, although it consistently exhibited the fastest inference speed among the local LLMs. Surprisingly, with some metrics (BLEURT and BERTScore), Mixtral (8x7B) even surpassed ChatGPT, and this effect was more pronounced on Ollama than on Llamafile.

### 5.2.3 GPT4ALL

The third platform tested was GPT4All, with the following results.

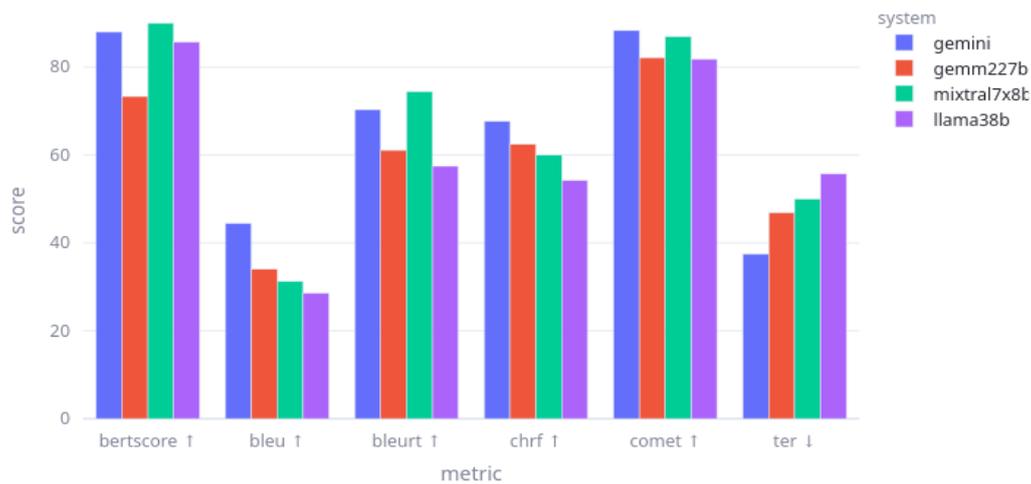

*Results with ChatGPT as reference*

| ChatGPT 4o mini | | | | | | | |
|---|---|---|---|---|---|---|---|
| **model** | **min.** | **bertscore** | **bleurt** | **comet** | **BLEU** | **chrF2** | **TER** |
| Gemini | 7 sec | 88,01 | 70,31 | 88,34 | 44,44 | 67,67 | 37,50 |
| Gemma2 27b | 8,08 | 73,34 | 61,09 | 82,16 | 34,01 | 62,44 | 46,88 |
| Mixtral 8x7b | 3,2 | 90,00 | 74,43 | 86,94 | 31,23 | 59,98 | 50,00 |
| Lama3 8b | 3,24 | 85,71 | 57,46 | 81,84 | 28,58 | 54,22 | 55,73 |

The results in the table confirm the online chatbots' significantly faster inference speed compared to the local models. Among the local models, Llama 3 and Mixtral were the fastest. Here, too, Mixtral (8x7B) outperformed Gemini according to BERTScore and BLEURT, but not according to BLEU and TER.

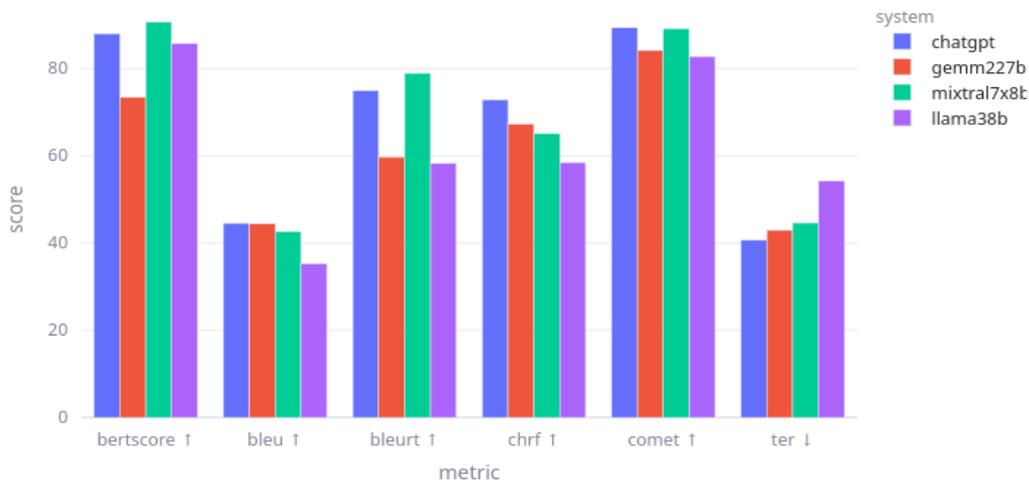

*Results with Gemini as reference*

**Gemini 1.5. Flash**

| model | min. | bertscore | bleurt | comet | BLEU | chrF2 | TER |
|---|---|---|---|---|---|---|---|
| ChatGPT | 12 sec | 88,01 | 74,94 | 89,36 | 44,57 | 72,85 | 40,68 |
| Gemma2 27b | 8,08 | 73,42 | 59,71 | 84,13 | 44,43 | 67,25 | 42,94 |
| Mixtral 8x7b | 3,2 | 90,69 | 78,89 | 89,16 | 42,63 | 65,1 | 44,63 |
| Lama3 8b | 3,24 | 85,76 | 58,25 | 82,78 | 35,27 | 58,47 | 54,24 |

The data clearly demonstrate that the online chatbots exhibit significantly lower latency. Among the local models, the smallest, Llama 3 (8B), and Mixtral proved to be the fastest. Overall, Gemma 2 (27B) proved to be the best-performing local model, although this performance came at the cost of increased latency.

## 5.3 Detailed translation task

For the second evaluation, the marketing text translation was performed using the following more detailed prompt:

> **You are an international marketing expert. Please translate this text from German into Italian. It is part of a website for a restaurant in the Alps near the city of Innsbruck; as such it should be used as an inviting and engaging version of the website for Italian-speaking tourists and attract guests to the restaurant:**

Again, the outputs from ChatGPT (version 4o Mini) and Gemini (version 1.5 Flash) served as references. These were compared against the translations produced by our three local models.

Due to the free interpretation and use of marketing-oriented language, the comparison should be interpreted cautiously. Consequently, we observed very low BLEU scores and very high TER scores.

## 5.3.1 Llamafile

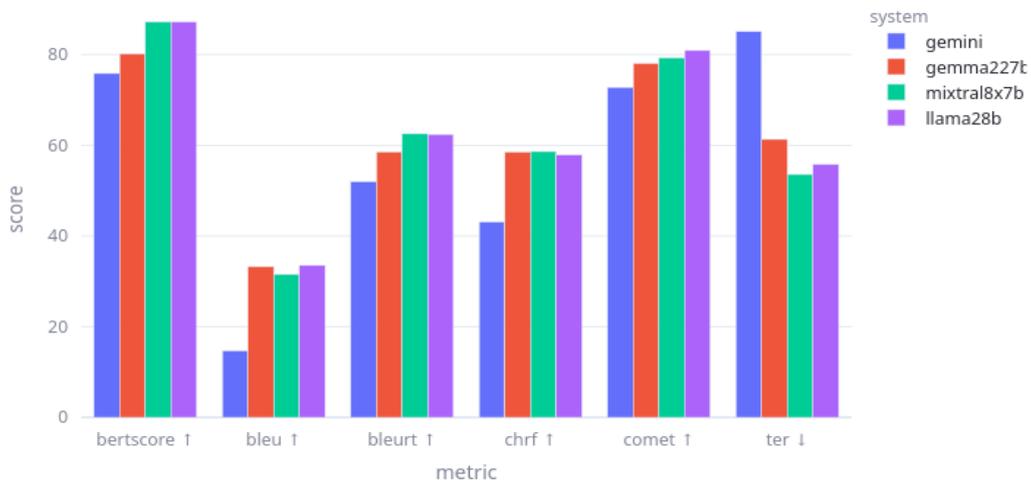

*Results with ChatGPT as reference*

| model | min. | bertscore | bleurt | comet | BLEU | chrF2 | TER |
|---|---|---|---|---|---|---|---|
| Gemini | 5,2 sec | 75,86 | 52 | 72,79 | 14,69 | 43,1 | 85,08 |
| Gemma2 27b | 6,25 | 80,2 | 58,54 | 78,06 | 33,23 | 58,51 | 61,33 |
| Mixtral 8x7b | 4,13 | 87,24 | 62,55 | 79,32 | 31,54 | 58,64 | 53,59 |
| Lama3 8b | 1,56 | 87,22 | 62,42 | 80,91 | 33,52 | 57,88 | 55,8 |

The data indicate that Gemini's translation differs significantly from the ChatGPT reference translation, while our local models' translations are more similar to the reference. This does not necessarily imply that Gemini performed poorly; it simply reflects notable differences in output.

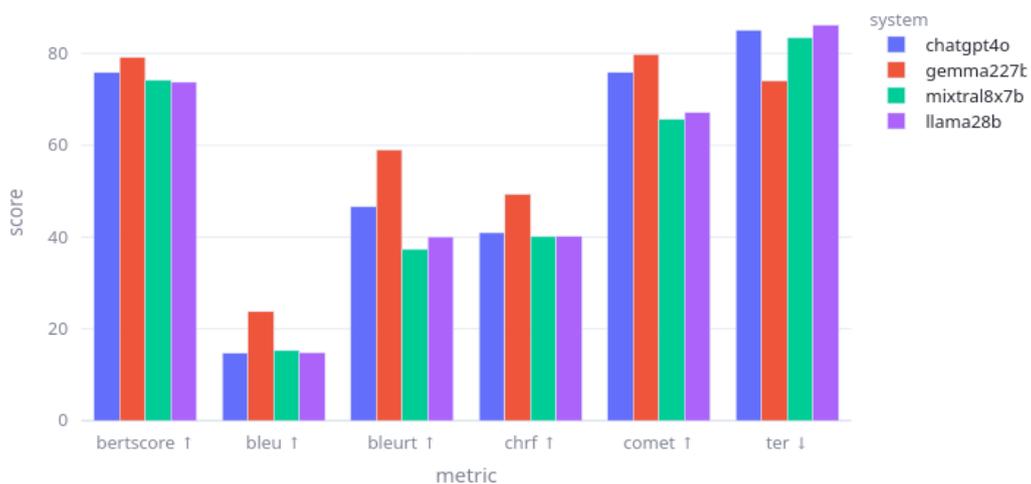

*Results with Gemini as reference*

Gemini 1.5. Flash

| model | min. | bertscore | bleurt | comet | BLEU | chrF2 | TER |
|---|---|---|---|---|---|---|---|
| ChatGPT | 9,42 sec | 75,86 | 46,63 | 75,87 | 14,69 | 40,96 | 85,08 |
| Gemma2 27b | 6,25 | 79,16 | 58,98 | 79,79 | 23,77 | 49,32 | 74,03 |
| Mixtral 8x7b | 4,13 | 74,2 | 37,38 | 65,66 | 15,28 | 40,11 | 83,43 |
| Lama3 8b | 1,56 | 73,75 | 40,04 | 67,15 | 14,82 | 40,15 | 86,19 |

The same observation applies to the comparison between ChatGPT and the Gemini reference translation. ChatGPT's BLEU score is lower than those of Gemma and Mixtral. However, this again simply highlights the differences in translation style. Interestingly, both Gemma 2 and Mixtral outperform ChatGPT in terms of the TER score as well.

### 5.3.2 Ollama

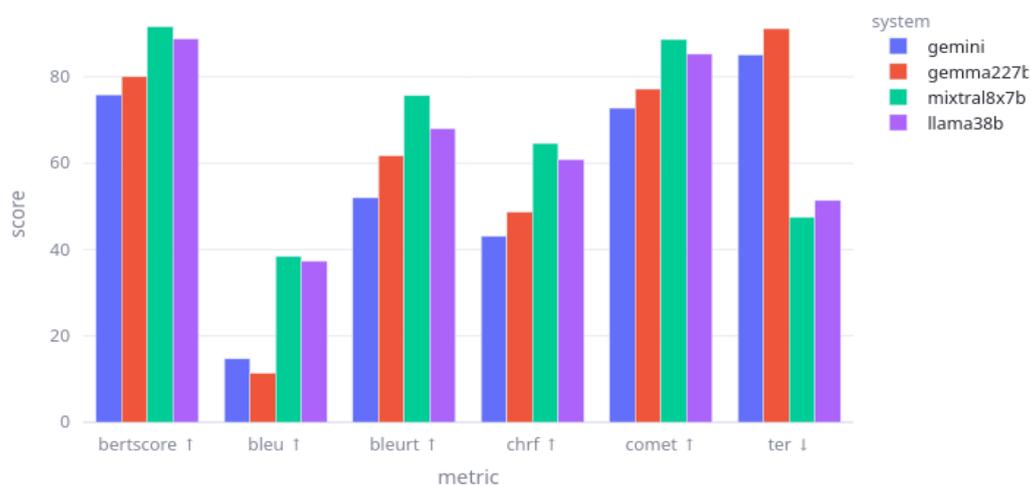

*Results with ChatGPT as reference*

ChatGPT 4o mini

| model | min. | bertscore | bleurt | comet | BLEU | chrF2 | TER |
|---|---|---|---|---|---|---|---|
| Gemini | 5,2 sec | 75,86 | 52 | 72,79 | 14,69 | 43,1 | 85,08 |
| Gemma2 27b | 6,37 | 80,07 | 61,72 | 77,18 | 11,35 | 48,7 | 91,16 |
| Mixtral 8x7b | 3,24 | 91,63 | 75,67 | 88,67 | 38,43 | 64,6 | 47,51 |
| Lama3 8b | 1,46 | 88,8 | 68,03 | 85,3 | 37,31 | 60,84 | 51,38 |

From the data we see that the translations by Mixtral and Llama are closer to ChatGPT than both Google based models which translated more freely.

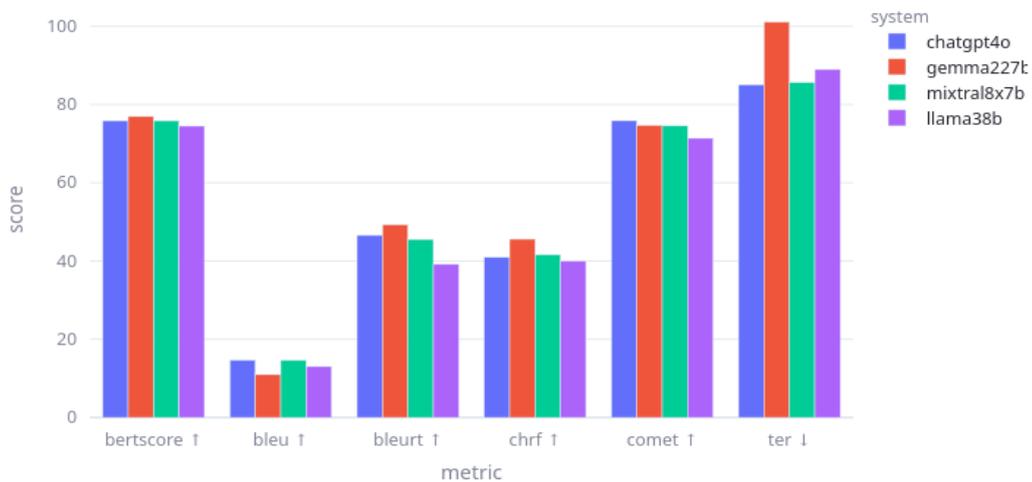

*Results with Gemini as reference*

| Gemini 1.5. Flash | | | | | | | |
|---|---|---|---|---|---|---|---|
| **model** | **min.** | **bertscore** | **bleurt** | **comet** | **BLEU** | **chrF2** | **TER** |
| ChatGPT | 9,42 sec | 75,86 | 46,63 | 75,87 | 14,69 | 40,96 | 85,08 |
| Gemma2 27b | 6,37 | 76,94 | 49,29 | 74,68 | 10,99 | 45,63 | 101,1 |
| Mixtral 8x7b | 3,24 | 75,85 | 45,55 | 74,56 | 14,67 | 41,6 | 85,64 |
| Lama3 8b | 1,46 | 74,5 | 39,17 | 71,41 | 13,05 | 40 | 88,95 |

Compared to Gemini, all models score poorly, exhibiting very low BLEU scores and very high TER scores. This is attributable to Gemini's highly interpretive and free formulation of the marketing text.

A TER score exceeding 100 indicates that the output requires more edits than the total number of words in the reference translation. Thus, in this instance, the Gemma 2 translation is substantially different from the Gemini translation.

### 5.3.3 GPT4All

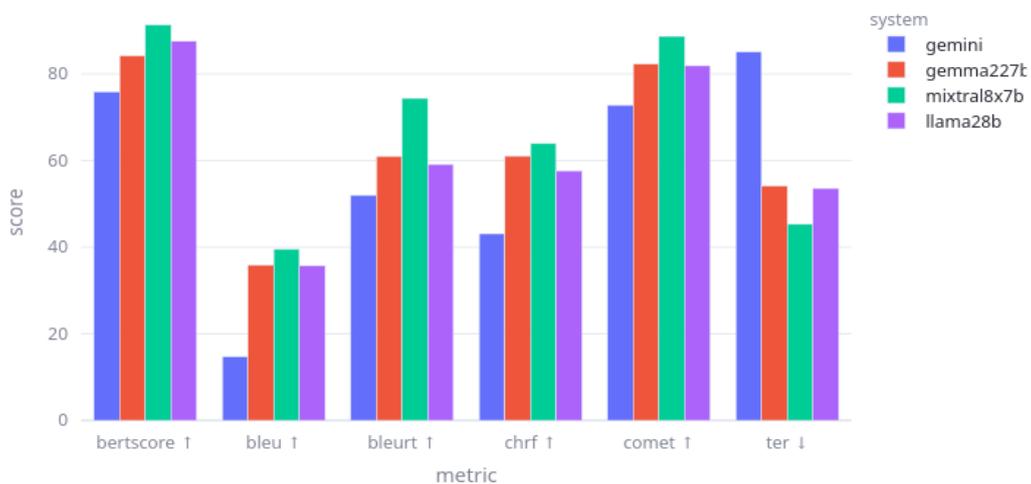

Results with ChatGPT as reference

ChatGPT 4o mini

| model | min. | bertscore | bleurt | comet | BLEU | chrF2 | TER |
|---|---|---|---|---|---|---|---|
| Gemini | 5,2 sec | 75,86 | 52 | 72,79 | 14,69 | 43,1 | 85,08 |
| Gemma2 27b | 9,57 | 84,19 | 60,97 | 82,29 | 35,88 | 61,05 | 54,14 |
| Mixtral 8x7b | 4,26 | 91,33 | 74,39 | 88,68 | 39,52 | 63,92 | 45,3 |
| Lama3 8b | 2,26 | 87,53 | 59,11 | 81,88 | 35,7 | 57,63 | 53,59 |

Again, a clear difference is observed between ChatGPT and Gemini, with Mixtral scoring higher than the other local models.

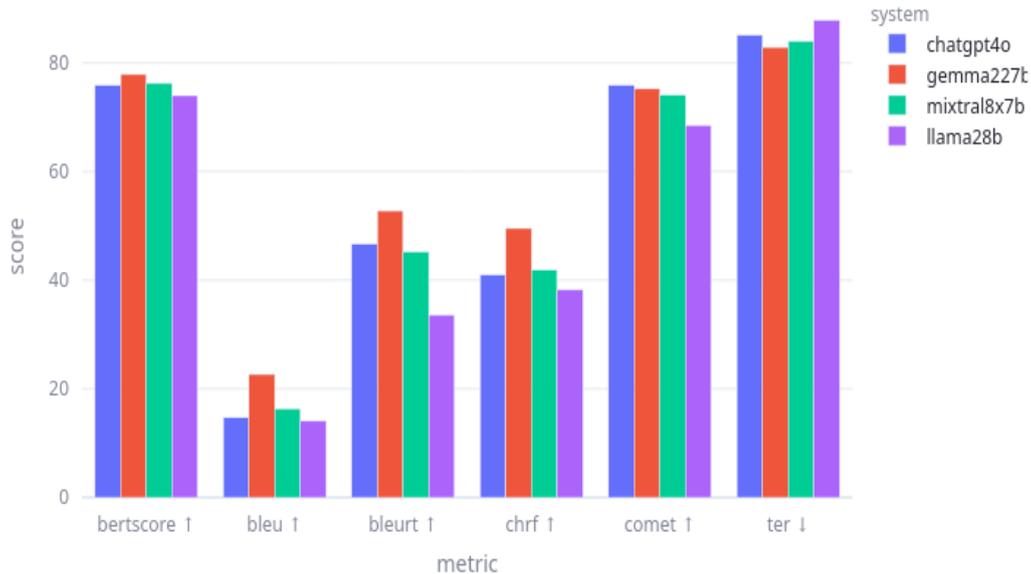

Results with Gemini as reference

Gemini 1.5. Flash

| model | min. | bertscore | bleurt | comet | BLEU | chrF2 | TER |
|---|---|---|---|---|---|---|---|
| ChatGPT | 9,42 sec | 75,86 | 46,63 | 75,87 | 14,69 | 40,96 | 85,08 |
| Gemma2 27b | 9,57 | 77,88 | 52,76 | 75,22 | 22,66 | 49,53 | 82,87 |
| Mixtral 8x7b | 4,26 | 76,27 | 45,21 | 74,06 | 16,33 | 41,87 | 83,98 |
| Lama3 8b | 2,26 | 73,98 | 33,58 | 68,48 | 14,07 | 38,23 | 87,85 |

As previously noted, Gemma 2 (27B) exhibits the greatest similarity to Gemini, although the overall scores remain low for all local models.

## 5.4 TMX Generation

The subsequent instruction, concerning sentence alignment and TMX conversion, was given immediately after the translation task, in a continuous interaction and without interrupting the chat session.

*Now, please align all the sentences of your Italian translation to the corresponding sentences of the German source text, sentence for sentence, and transform it into a translation memory in the TMX format. Please adhere to the format in this example:*
```
<?XML version="1.0" encoding="UTF-8"?>
<!DOCTYPE tax SYSTEM "tmx11.dtd">
<tmx version="1.1">
<header adminlang="EN-US" datatype="plaintext" segtype="sentence" srclang="it"/>
<body>
<tu>
<tuv lang="it">
<seg>Italian text</seg>
```

```
</tuv>
<tuv lang="de-AT">
<seg>German text</seg>
</tuv>
</tu>
</body>
</tmx>
```

### 5.4.1 LLamafile

This is a summary of the performance of three large language models (LLMs) on another preparatory translation task, specifically their ability to generate Translation Memory Exchange (TMX) files. Here's a breakdown of the issues encountered on the platform Llamafile.

The model Llama3 (8B) failed to process the entire task, halting after only three segments. Even after being prompted to continue, it could not complete the process, resulting in an unusable TMX file.

Mixtral (8x7B) also struggled with TMX generation. It stopped after processing only three translation units in the first attempt. A second attempt resulted in mixed languages and the insertion of an English equivalent, indicating potential confusion with the task. Even with explicit instructions and the same prompt, the model failed again, leading to an unusable TMX file.

The LLM Gemma2 (27B) failed to produce a TMX file altogether, stopping abruptly after receiving the input, indicating a more fundamental issue with processing the request.

In essence, all three models as Llamafiles encountered significant challenges in generating usable TMX files, highlighting potential limitations in their ability to handle this specific format or task.

| Llamafile | time | n° of TU |
|---|---|---|
| Gemma2 (27b) | - | - |
| Mixtral 8x7b) | 04:13,43 | 9 |
| Llama3 (8b) | 01:34,81 | 12 |

### 5.4.2 Ollama

On Ollama Llama3 (8B) required 4 minutes and 37.23 seconds to generate 12 translation units. However, the resulting TMX file lacked all opening <TU> tags, rendering the translation units unusable. While this specific error, a consistent omission, makes the TMX file unusable, it could be readily corrected.

Mixtral's initial attempt resulted in hallucinations, with the model introducing sentences absent from the source text alongside the original content. Following a restart of the chat session, Mixtral generated 9 translation units in 6 minutes and 56.21 seconds, but terminated prematurely within a word ("La nostra veranda storic") in one of the final segments.

Gemma2 (27B) took significantly longer, 17 minutes and 35 seconds, to produce 12 translation units. The resulting TMX file contained a single, albeit minor, error: a duplicated closing </TU> tag at the end of the file.

| Ollama | time | n° of TU |
|---|---|---|
| Gemma2 (27b) | 17:34,71 | 12 |
| Mixtral 8x7b) | 06:56,21 | 9 |
| Llama3 (8b) | 04:37,23 | 12 |

### 5.4.3 GPT4All

Llama3 (8B), running on GPT4All, generated six translation units in 5 minutes and 24.33 seconds, producing a TMX file with the correct format. However, the output was incomplete, containing only one item from the list of offerings of the restaurant, and merging two sentences into a single segment.

Mixtral (8x7B) generated eight translation units in 11 minutes and 33.37 seconds. The alignment was flawed,

combining two sentences into one segment on two occasions, which reduced the total number of segments.

Gemma2 (27B) produced 13 translation units in 49 minutes and 3.67 seconds. While the TMX format was correct, the output exhibited some issues towards the end of the text. Specifically, one sentence was split into two segments, and part of the final sentence was omitted. The extended processing time suggests that manual TMX generation might be more efficient in this scenario. This performance can likely be attributed to the limitations of the CPU-based hardware used. A more powerful setup with one or more GPUs could potentially improve processing time and overall performance.

| GPT4All | time | n° of TU |
|---|---|---|
| Gemma2 (27b) | 49:03,67 | 13 |
| Mixtral 8x7b) | 11:33,37 | 8 |
| Llama3 (8b) | 05:24,33 | 6 |

The generated TMX file was tested in the translation management system OmegaT, which segmented the source text into ten units. Due to alignment inconsistencies, not all translations were displayed correctly, and a 100% match score was achieved only sporadically. Among the models tested, Gemma required the least amount of post-processing to create a functional translation memory. While Ollama exhibited faster processing times than GPT4All, it encountered the most significant issues with the TMX file format. Consistent with previous observations, Llama3 (8B) was the fastest model but produced the lowest quality output.

For pre-translation tasks like this, extended processing times can be mitigated by scheduling jobs overnight or utilizing a dedicated machine to prevent disruptions to other ongoing work.

## 5.5 Terminology extraction

Compiling a glossary of terms from a specialized text corpus is another common preparatory task for translation. This can be achieved either by searching the web for relevant texts or by leveraging a provided local corpus. Using web searches raises confidentiality concerns and conflicts with the principle of using local models. The latter necessitates that AI platforms possess the capability to integrate and retrieve information from local text corpora, by uploading texts or by Retrieval Augmented Generation (RAG). Llamafile lacks the functionality of uploading files and RAG, limiting our evaluation to two platforms.

Following the integration and compilation of the small text corpus, we prompted the system as follows:

Please compile a German Italian bilingual list with two columns of at least 50 terms related to the bylaws for Austrian and Italian limited liability companies in the knowledge corpus. Please extract only terms from the knowledge corpus. Terms should be specific to this type of text and domain.

### 5.5.1 Ollama

In Ollama, the texts were uploaded via Settings / Manage Knowledge, creating two collections: Austrian statutes and Italian statutes, each containing ten documents. The embedding and vectorization process required approximately six minutes using the "nomic-embed-text-V1.5" model.

Within Ollama PageAssist's RAG settings, the "Websearch" option was disabled, and the "Knowledge" option was enabled. However, only a single knowledge collection could be selected at a time. Consequently, the German and Italian corpora were combined into a single collection before submitting the prompt. This approach proved unsuccessful; only the German portion of the corpus was cited as a source. Italian terms were not extracted, and the system instead invented terms, occasionally incorporating English terms and exhibiting signs of hallucination.

Ideally, an aligned corpus, such as a translation memory (TMX) file, would be required. However, if a TMX file is available, compiling a term list might offer limited additional value for translation. Furthermore, aligning original legal texts would be a complex undertaking due to substantial differences arising from distinct legal frameworks.

Therefore, an alternative three-step approach was adopted: First, two separate knowledge corpora were embedded. Second, the most frequent terms were extracted from each language corpus independently. Finally, these terms were compiled into a bilingual glossary. We used the following prompts:

> Please compile a list of the 30 most frequent German/Italian special terms from the knowledge corpus.

> Please combine the German list and the Italian list of terms into a bilingual term list

Llama3 (8B) struggled with this task, producing a limited output of only 12 bilingual terms. The generated terms were often incorrect, mixed English and German, and included invented terms.

Mixtral (8x7B), using the combined knowledge resource, initially generated a list of 18 bilingual terms in 6 minutes and 50.43 seconds. However, these terms primarily consisted of phrases and multi-word combinations. Subsequently, after processing the German and Italian knowledge resources separately, Mixtral (8x7B) created two lists of 30 monolingual terms each. These were then combined to form a bilingual list containing five terms from each language, each accompanied by two example sentences.

| Ollama | time German list | n° of monolingual terms | time Italian list | n° of monolingual terms | time combined list | n° of bilingual terms |
|---|---|---|---|---|---|---|
| Gemma2 (27b) | - | - | - | - | 03:44,39 | 18 |
| Mixtral 8x7b) | 06:28,46 | 30 | 06:25,69 | 30 | 12:32,80 | 5 |
| Llama3 (8b) | 02:25,51 | 30 | 02:31,15 | 30 | 02:19,23 | 12 |

Gemma2 (27B) again demonstrated the strongest performance on the initial prompt using the combined corpus, generating 18 accurate bilingual terms in 6 minutes and 48.40 seconds. However, subsequent attempts to expand this list were unsuccessful. Furthermore, Gemma explicitly declined to process the second approach, which involved separate monolingual corpora, citing limitations in its capabilities. Specifically, it stated, "I cannot perform complex tasks like frequency analysis to determine the 30 most frequent special terms" and "I do not have the capabilities for such advanced linguistic processing".

### 5.5.2 GPT4All

The GPT4All platform implements RAG through its LocalDocs option. This option allows for the integration of one or more document collections. When prompting the system, users can select which collection(s) to include in their request. Again, we used the "nomic-embed-text-V1.5" embedding model and uploaded both German and Italian text files.

Using Llama3 (8B) within GPT4All, a list of 24 bilingual terms was generated in 4 minutes and 49.81 seconds. While the result was not unsatisfactory, it lacked specificity for the domain of bylaws. With the LocalDoc collection containing only German texts, Llama3 (8B) generated a list of the 30 most frequent German terms in 2 minutes and 24.11 seconds. Again, the system failed to recognize the specific domain and returned terms primarily related to a single limited liability company (in this case, a university). The Italian monolingual list took 5 minutes and 32.17 seconds to generate. Combining the two lists into a bilingual term list required 3 minutes and 37.74 seconds. The output was acceptable, but while the model explicitly stated, "Here is the combined bilingual list of terms in German and Italian", Italian terms were enclosed in parentheses, and English terms were listed as equivalents, as in "Gesellschaft (Società) – Company".

| GPT4All | time German list | n° of monolingual terms | time Italian list | n° of monolingual terms | time combined list | n° of bilingual terms |
|---|---|---|---|---|---|---|
| Gemma2 (27b) | 18:09,14 | 30 | 15:14,12 | 30 | 10:57,12 | 19 |
| Mixtral 8x7b) | 06:36,57 | 30 | 05:39,25 | 30 | 07:50,28 | 30 |
| Llama3 (8b) | 02:24,11 | 30 | 05:32,17 | 30 | 03:37,74 | 30 |

With our initial prompt, Mixtral (8x7B) generated 39 bilingual terms in 12 minutes and 56.21 seconds. The resulting glossary was generally good, but contained some incorrect equivalences, and all entries included English equivalents. Despite this, the system stated, "Response: German (Translation of Terms in the Text) | Italian (Original Terms in the Text)", even though the LocalDocs corpus contained both languages, meaning all terms should have been original. Repeating the same prompt with the addition of "without English terms" yielded a better glossary, but still with a few incorrect equivalences. This time, the output began with, "Response: | Italiano (Bylaws for Italian SRL) | Deutsch (Bylaws für GmbH in Österreich)". Changing the procedure and initially requesting monolingual terms resulted in 30 Italian terms in 5 minutes and 39.25 seconds, and 30 German terms in 6 minutes and 36.57 seconds. Combining these lists into a bilingual list took 7 minutes and 50.28 seconds.

Our initial attempt with Gemma2 (27B) was rejected with the same "more context and processing power" message encountered with Ollama. A second attempt, focusing on monolingual lists, required 18 minutes and 9.14 seconds for the German list and 15 minutes and 14.12 seconds for the Italian list. Interestingly, the terms were grouped into very specific subdomains related to limited liability companies, although the category names differed between the Italian and German lists. Combining these into a bilingual term list took 10 minutes and 57.12 seconds and resulted in 19 term pairs. Remarkably, this combined list included some terms absent from both the original German and Italian lists. While the output was acceptable, it lacked comprehensiveness. Overall, Gemma2 (27B) produced the best results, but at the cost of extremely long processing times. However, this demonstrates that Gemma2 (27B) in conjunction with GPT4All is, in principle, capable of performing this task satisfactorily. With more powerful hardware, this approach would be feasible.

In summary, terminology extraction proved generally unsatisfactory across all tested platforms and models. Results were often either of poor quality or lacked comprehensiveness, and processing times were excessively long. This inadequacy in term extraction is confirmed by Heinisch (2024) even for online LLMs: "limitations of LLMs already in this step" (Heinisch 2024: 37). Based on our experiments, fully automating terminology extraction is not currently feasible; it consistently requires human oversight and quality control of the output. It is important to note that more specialized and higher-performing tools exist for this specific task.

## 5.6 TM-Workflow

### *5.6.1 Sentence translation with simple prompt*

Presented with a simple prompt the LLM translated the sample sentence in the way a machine translation plugin would do.

| **Please translate this text from German into Italian** |
| --- |

**Llamafile**

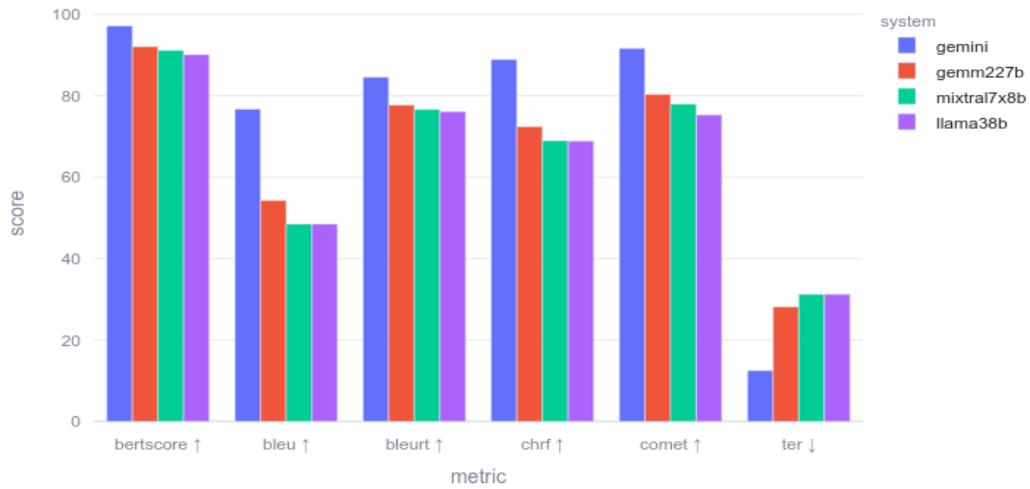

*Results with ChatGPT as reference*

| simple prompt Llamafile | | | | | | |
|---|---|---|---|---|---|---|
| model | time sec. | bertscore | bleurt | comet | BLEU | chrF2 | TER |
| Gemini | 3 | 97,2 | 84,6 | 91,6 | 76,7 | 88,9 | 12,5 |
| Gemma2 (27b) | 74 | 92,1 | 77,7 | 80,4 | 54,3 | 72,4 | 28,1 |
| Mixtral 8x7b) | 54 | 91,1 | 76,6 | 78,0 | 48,5 | 68,9 | 31,3 |
| Llama3 (8b) | 24 | 90,1 | 76,1 | 75,3 | 48,5 | 68,9 | 31,3 |

Results from Mateo indicated the translation by Gemini as the nearest to the output from ChatGPT. Among the local models scores were almost identical with Gemma2 (27B) slightly better than the others.

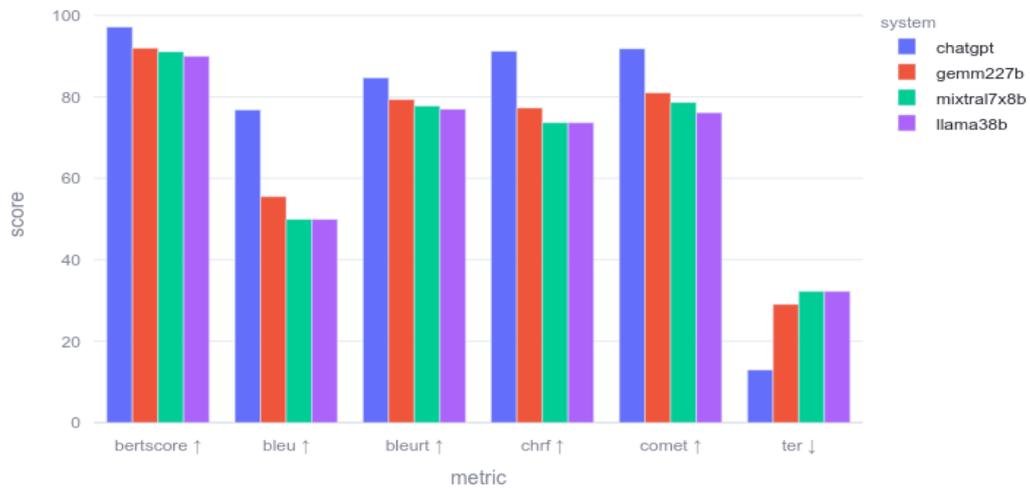

*Results with Gemini as reference*

| simple prompt Llamafile | | | | | | | |
|---|---|---|---|---|---|---|---|
| model | time sec. | bertscore | bleurt | comet | BLEU | chrF2 | TER |
| ChatGPT | 2 | 97,2 | 84,7 | 91,8 | 76,8 | 91,2 | 12,9 |
| Gemma2 (27b) | 74 | 92,0 | 79,4 | 81,0 | 55,5 | 77,3 | 29,0 |
| Mixtral (8x7b) | 54 | 91,1 | 77,7 | 78,6 | 49,9 | 73,7 | 32,3 |
| Llama3 (8b) | 24 | 90,0 | 77,0 | 76,1 | 49,9 | 73,6 | 32,3 |

Taking Gemini as a reference we were presented with the same results: ChatGPT is nearest to the reference and Gemma 2 (27B) with a slight advantage over the other local LLMs.

**Ollama**

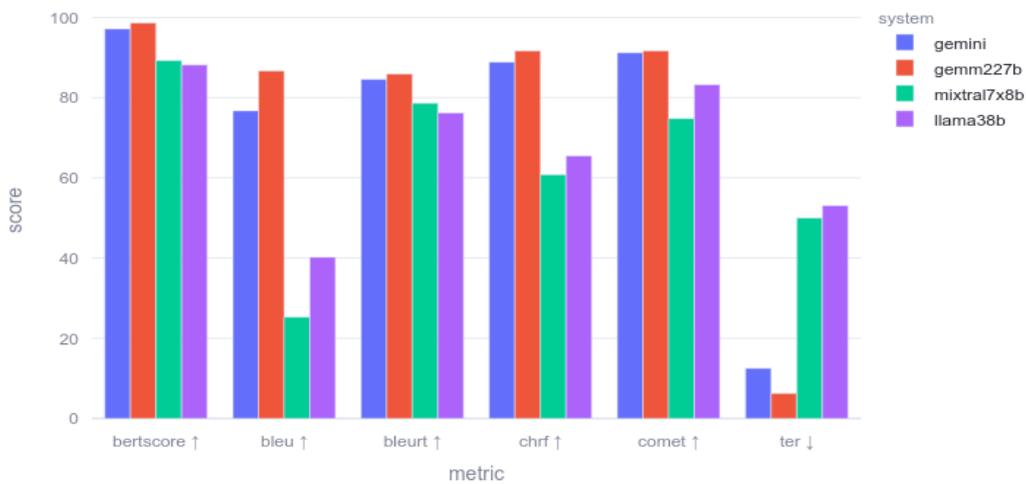

*Results with ChatGPT as reference*

| simple prompt Ollama | | | | | | | |
|---|---|---|---|---|---|---|---|
| model | time sec. | bertscore | bleurt | comet | BLEU | chrF2 | TER |
| Gemini | 3 | 97,2 | 84,6 | 91,2 | 76,7 | 88,9 | 12,5 |
| Gemma2 (27b) | 64 | 98,7 | 85,9 | 91,7 | 86,7 | 91,7 | 6,3 |
| Mixtral 8x7b) | 53 | 89,3 | 78,6 | 74,8 | 25,3 | 60,8 | 50,0 |
| Llama3 (8b) | 24 | 88,2 | 76,2 | 83,3 | 40,2 | 65,5 | 53,1 |

Again Gemma2 (27B) was almost identical to Gemini, even more than ChatGPT, scoring a BLEU of 86,7 and a TER of 6,3. Interestingly, Llama3 (8B) under Ollama performed better than Mixtral with a BLEU score of 40,2 against Mixtrals's 25,3, which is in contrast with the higher TER for Llama3 (8B).

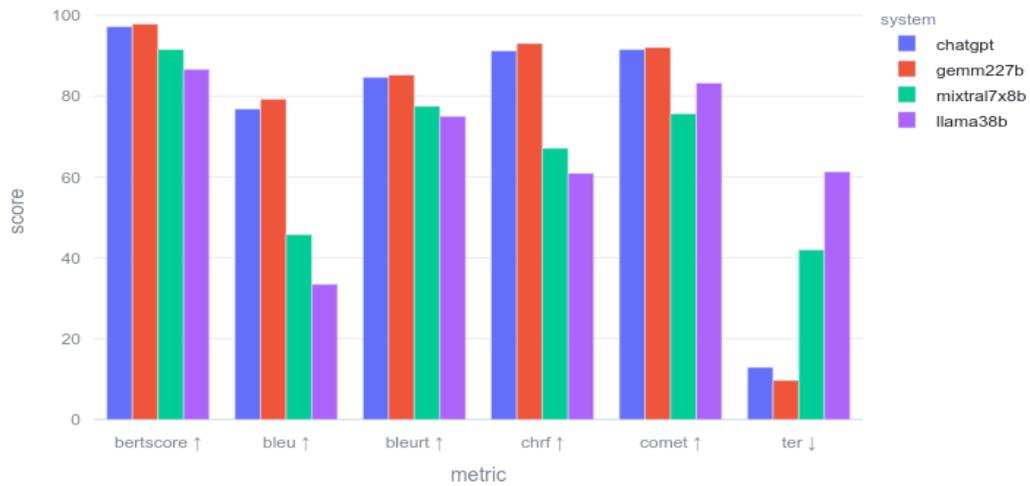

*Results with Gemini as reference*

simple prompt Ollama

| model | time sec. | bertscore | bleurt | comet | BLEU | chrF2 | TER |
|---|---|---|---|---|---|---|---|
| ChatGPT | 3 | 97,2 | 84,7 | 91,5 | 76,8 | 91,2 | 12,9 |
| Gemma2 (27b) | 64 | 97,9 | 85,2 | 92,1 | 79,2 | 93,0 | 9,7 |
| Mixtral 8x7b) | 53 | 91,5 | 77,5 | 75,6 | 45,7 | 67,1 | 41,9 |
| Llama3 (8b) | 24 | 86,6 | 75,0 | 83,3 | 33,5 | 60,9 | 61,3 |

**GPT4All**

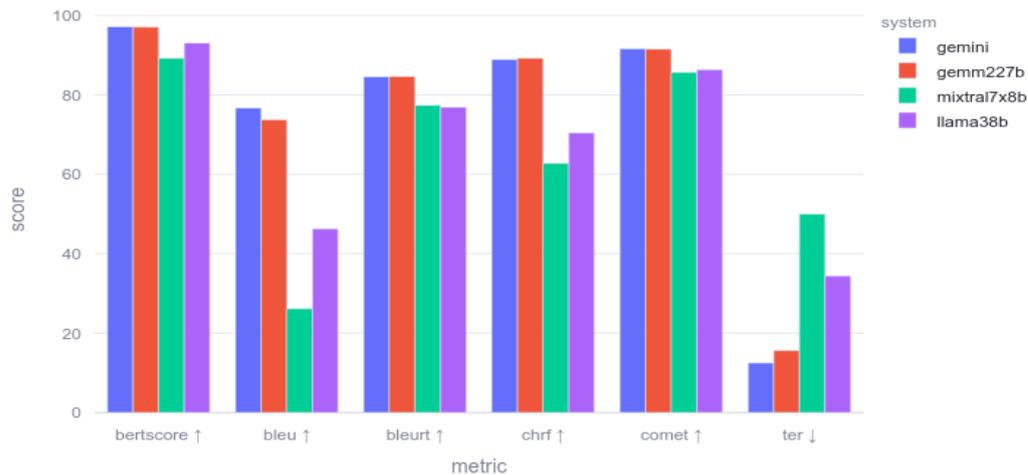

*Results with ChatGPT as reference*

| simple prompt GPT4All | | | | | | |
|---|---|---|---|---|---|---|
| model | time sec. | bertscore | bleurt | comet | BLEU | chrF2 | TER |
| Gemini | 3 | 97,2 | 84,6 | 91,6 | 76,7 | 88,9 | 12,5 |
| Gemma2 (27b) | 64 | 97,1 | 84,7 | 91,6 | 73,7 | 89,3 | 15,6 |
| Mixtral (8x7b) | 53 | 89,2 | 77,4 | 85,7 | 26,2 | 62,7 | 50,0 |
| Llama3 (8b) | 24 | 93,1 | 76,9 | 86,3 | 46,2 | 70,4 | 34,4 |

Among the local models Gemma2 (27B) scores best. Exactly as under Ollama, Llama3 (8B) has a higher BLEU and a lower TER than Mixtral (8x7B).

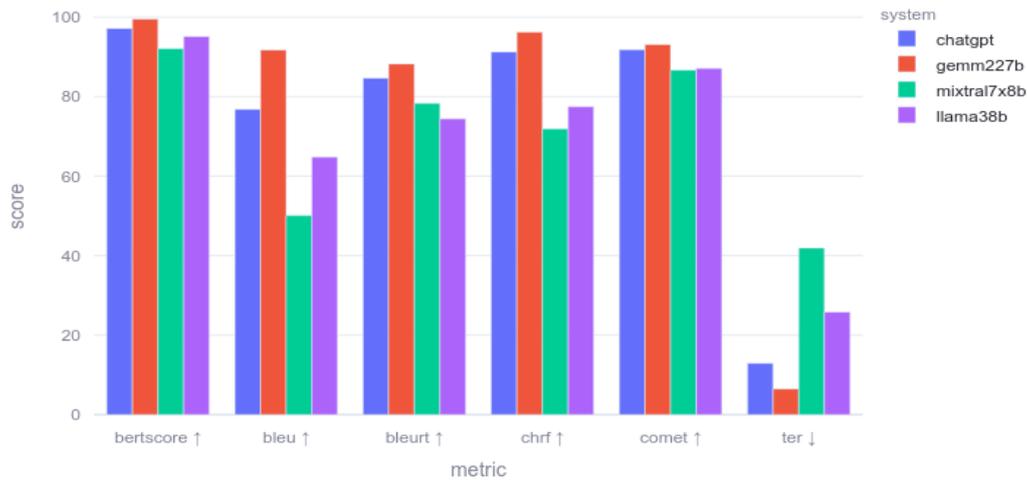

*Results with Gemini as reference*

| simple prompt GTP4All | | | | | | |
|---|---|---|---|---|---|---|
| model | time sec. | bertscore | bleurt | comet | BLEU | chrF2 | TER |
| ChatGPT | 2 | 97,2 | 84,7 | 91,8 | 76,8 | 91,2 | 12,9 |
| Gemma2 (27b) | 82 | 99,5 | 88,2 | 93,1 | 91,7 | 96,2 | 6,5 |
| Mixtral 8x7b) | 64 | 92,1 | 78,3 | 86,7 | 50,1 | 71,9 | 41,9 |
| Llama3 (8b) | 30 | 95,1 | 74,4 | 87,1 | 64,8 | 77,4 | 25,8 |

Again Llama3 (8B) has a higher BLEU than Mixtral. Subjectively, the translation output of Mixtral (8x7B) would be valued higher than the one from Llama3 (8B): this shows that an assessment on the basis of a single sentence, where a few corresponding words are decisive, is rather questionable.

### 5.6.2 Sentence translation with a detailed prompt

We expected better results with a more detailed prompt with the indication of persona, domain, legal background and scope of the translation.

**You are a company lawyer. Please translate this text from Italian into German. Do not give any explanations and try to be as quick as possible. The text is an extract from the bylaws of an Italian limited liabilities company. The translation should be used as a documentation for an Austrian company.**

**Llamafile**

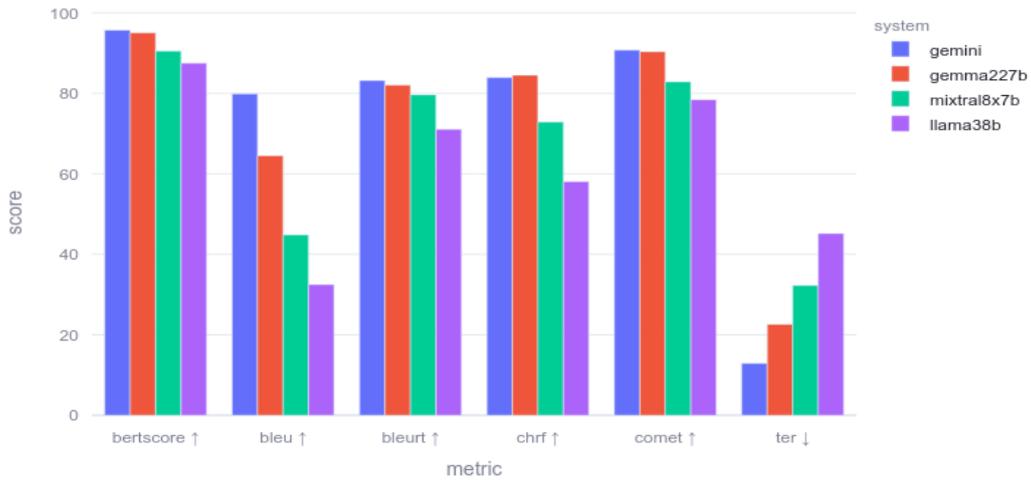

*Results with ChatGPT as reference*

| detailed prompt Llamafile | | | | | | |
|---|---|---|---|---|---|---|
| model | time sec. | bertscore | bleurt | comet | BLEU | chrF2 | TER |
| Gemini | 2 | 95,7 | 83,2 | 90,7 | 79,9 | 83,9 | 12,9 |
| Gemma2 (27b) | 52 | 95,1 | 82,0 | 90,4 | 64,5 | 84,5 | 22,6 |
| Mixtral 8x7b) | 55 | 90,5 | 79,6 | 82,9 | 44,8 | 72,9 | 32,3 |
| Llama3 (8b) | 19 | 87,5 | 71,1 | 78,4 | 32,5 | 58,1 | 45,2 |

Taking the output from Gemini as the reference translation we got the following results.

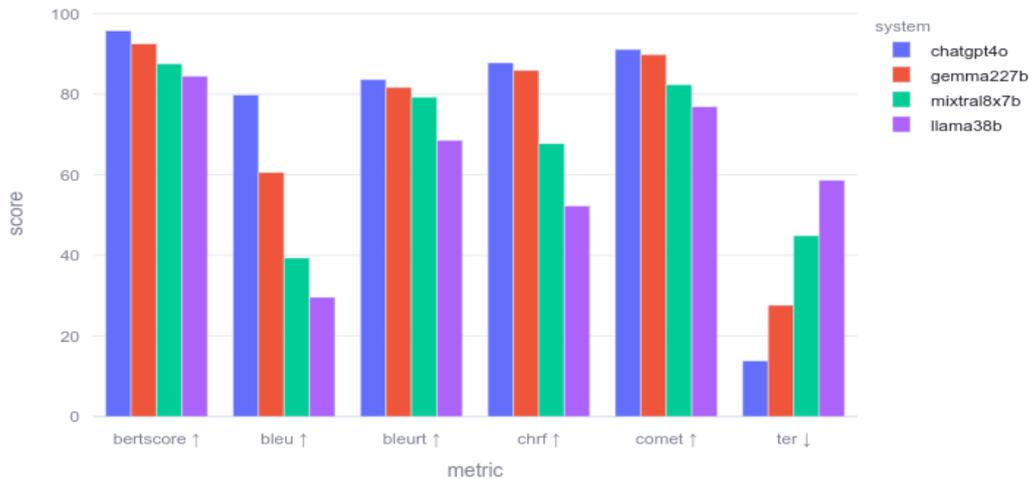

*Results with Gemini as reference*

| detailed prompt Llamafile | | | | | | |
|---|---|---|---|---|---|---|
| model | time sec. | bertscore | bleurt | comet | BLEU | chrF2 | TER |
| ChatGPT | 5 | 95,7 | 83,6 | 91,1 | 79,8 | 87,8 | 13,8 |
| Gemma2 (27b) | 52 | 92,5 | 81,6 | 89,8 | 60,6 | 85,9 | 27,6 |
| Mixtral 8x7b) | 55 | 87,5 | 79,3 | 82,3 | 39,4 | 67,7 | 44,8 |
| Llama3 (8b) | 19 | 84,5 | 68,5 | 76,9 | 29,6 | 52,3 | 58,6 |

**Ollama**

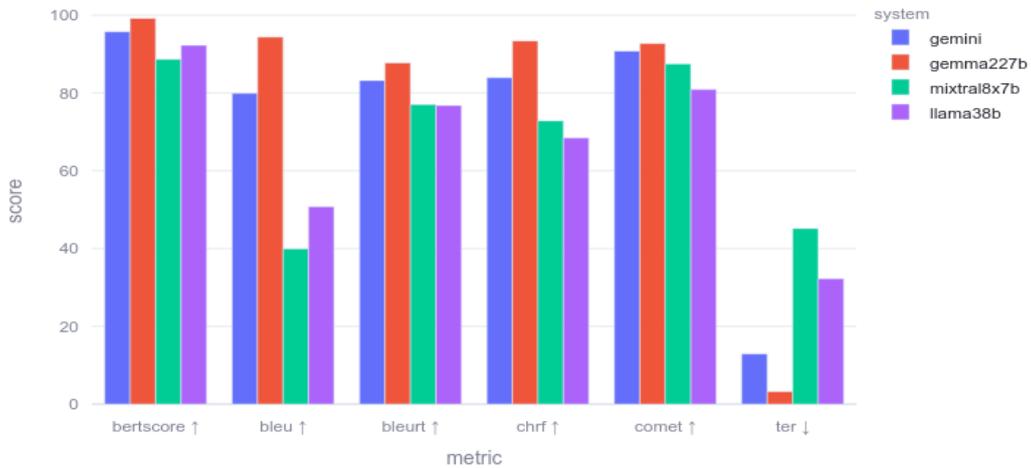

*Results with ChatGPT as reference*

| detailed prompt Ollama | | | | | | | |
|---|---|---|---|---|---|---|---|
| model | time sec. | bertscore | bleurt | comet | BLEU | chrF2 | TER |
| Gemini | 5 | 95,7 | 83,2 | 90,7 | 79,9 | 83,9 | 12,9 |
| Gemma2 (27b) | 51 | 99,2 | 87,8 | 92,7 | 94,4 | 93,4 | 3,2 |
| Mixtral 8x7b) | 43 | 88,6 | 77,0 | 87,4 | 39,9 | 72,9 | 45,2 |
| Llama3 (8b) | 22 | 92,2 | 76,8 | 80,9 | 50,8 | 68,4 | 32,3 |

Translations from ChatGPT and Gemma2 (27B) are almost identical, only the last term is different. Gemma uses *Stammkapital*, which is the specific term, while ChatGPT uses *Gesellschaftskapital*, which is not wrong, but a more general term. This explains the TER value of 3,2.

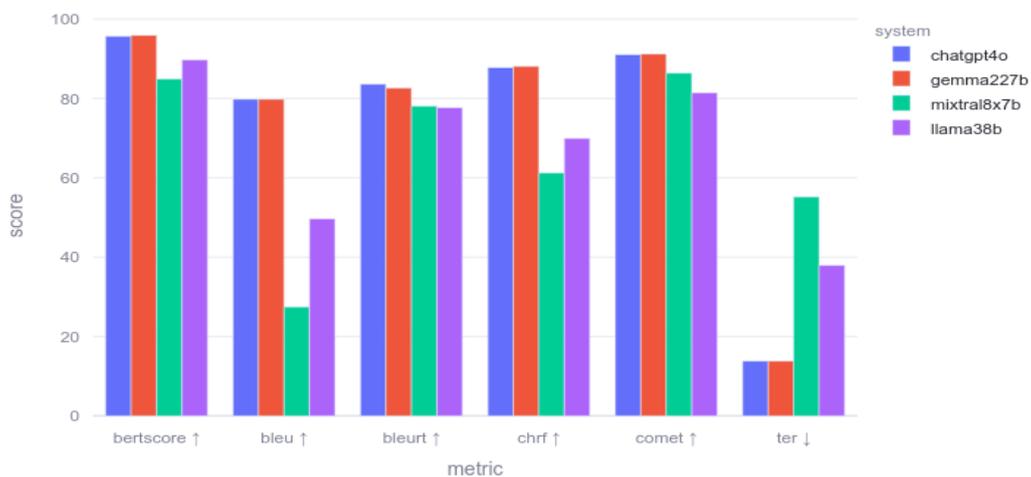

*Results with Gemini as reference*

| detailed prompt Ollama | | | | | | | |
|---|---|---|---|---|---|---|---|
| model | time sec. | bertscore | bleurt | comet | BLEU | chrF2 | TER |
| ChatGPT | 2 | 95,7 | 83,6 | 91,1 | 79,8 | 87,8 | 13,8 |
| Gemma2 (27b) | 51 | 95,9 | 82,6 | 91,2 | 79,8 | 88,1 | 13,8 |
| Mixtral 8x7b) | 43 | 84,9 | 78,1 | 86,4 | 27,4 | 61,2 | 55,2 |
| Llama3 (8b) | 22 | 89,7 | 77,7 | 81,4 | 49,7 | 69,9 | 37,9 |

**GPT4All**

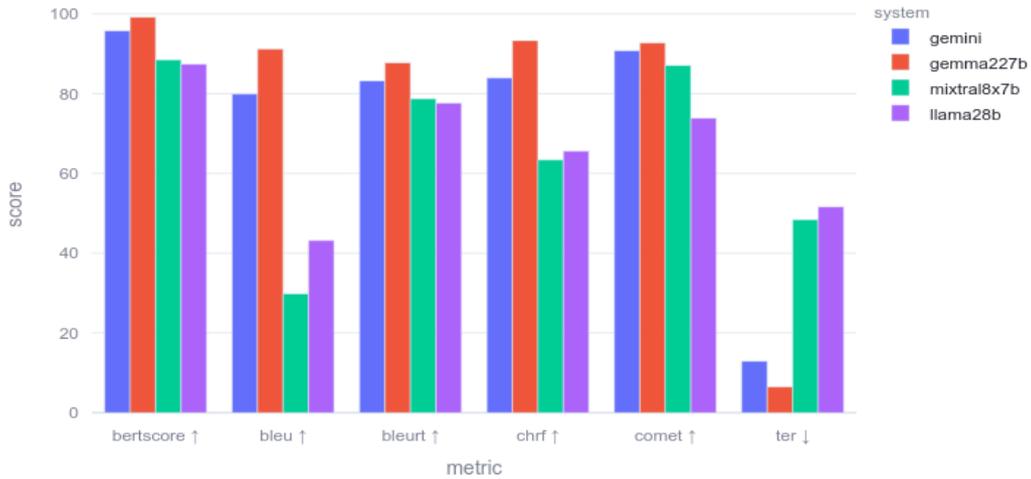

*Results with ChatGPT as reference*

| detailed prompt GPT4All | | | | | | | |
|---|---|---|---|---|---|---|---|
| model | time sec. | bertscore | bleurt | comet | BLEU | chrF2 | TER |
| Gemini | 3 | 95,7 | 83,2 | 90,7 | 79,9 | 83,9 | 12,9 |
| Gemma2 (27b) | 92 | 99,2 | 87,8 | 92,7 | 91,2 | 93,3 | 6,5 |
| Mixtral 8x7b) | 62 | 88,5 | 78,8 | 87,1 | 29,8 | 63,4 | 48,4 |
| Llama3 (8b) | 25 | 87,4 | 77,6 | 73,9 | 43,2 | 65,6 | 51,6 |

Gemma2 (27B) is nearer to the results of ChatGPT than Gemini which tends to deliver a more free translation. Llama3 (8B) has a higher BLEU score than Mixtral, meaning the translation is nearer to the one by ChatGPT, but strangely also a higher TER score, meaning more edits have to be performed to adapt its translation to the one delivered by ChatGPT.

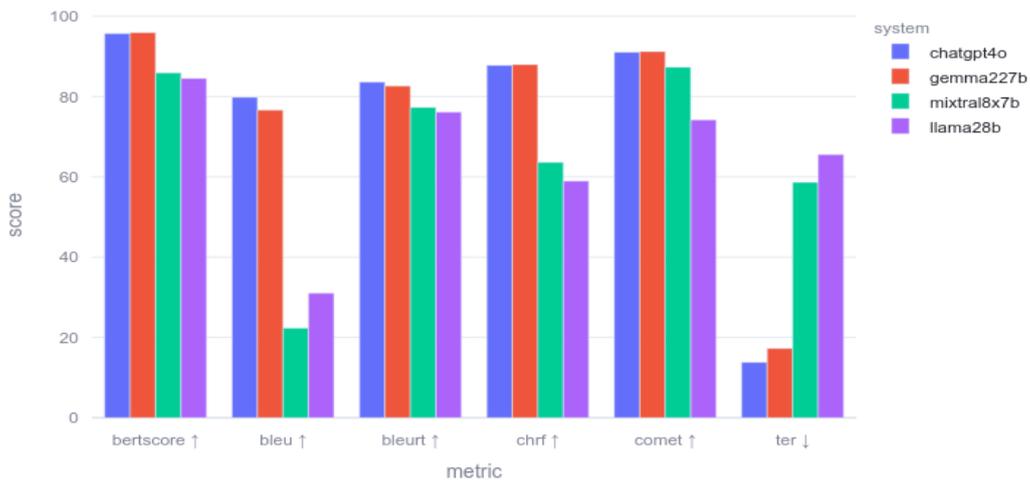

| detailed prompt GPT4All | | | | | | | |
|---|---|---|---|---|---|---|---|
| model | time sec. | bertscore | bleurt | comet | BLEU | chrF2 | TER |
| ChatGPT | 3 | 95,7 | 83,6 | 91,1 | 79,8 | 87,8 | 13,8 |
| Gemma2 (27b) | 92 | 95,9 | 82,6 | 91,2 | 76,6 | 88,0 | 17,2 |
| Mixtral 8x7b) | 62 | 85,9 | 77,3 | 87,3 | 22,3 | 63,6 | 58,6 |
| Llama3 (8b) | 25 | 84,5 | 76,1 | 74,2 | 31,0 | 58,9 | 65,5 |

Using Gemini as a benchmark, we observed contrasting results with ChatGPT, which achieved a lower TER score than Gemma 2 (27B) and a higher BLEU score. At the lower end of the performance spectrum, Llama 3 (8B) and Mixtral (8x7B) exhibited similar behavior, with Llama 3 (8B) surpassing Mixtral (8x7B) in BLEU score but lagging behind in TER score.

# 6 Conclusion

This study investigated the applicability of open-source, locally hosted Large Language Models (LLMs) within a standard freelance translation workflow. While acknowledging the broader benefits of open-source generative AI (Eiras et al., 2024), our focus was on the practical challenges and opportunities presented by local LLM deployment. Our findings suggest that current local LLM systems, when run on standard desktop hardware, are not yet mature enough to fully replace or seamlessly integrate with existing professional translation tools. They do not yet match the quality and, especially, the latency of commercial online AI solutions. However, we should point out that larger local models and high-grade GPU hardware are crucial for achieving higher quality results, pointing towards a promising future as hardware capabilities advance. Indeed, the computational demands of local LLMs are pushing traditional computing architectures to their limits, highlighting the need for specialized hardware like AI accelerators and neuromorphic chips.

The evaluation of translation quality presents a complex picture: summing BLEU scores for individual LLMs (text/sentence translation) yields 409/908 for Gemma2 (27B), 383/449 for Mixtral (8x7B), and 338/520 for Llama3 (8B). Among the models tested, Gemma2 (27B) demonstrated the best balance between accuracy and output quality, consistent with the findings of Cui et al. (2025). Mixtral (8x7B), despite its larger size and due to its Mixture-of-Experts architecture, exhibited lower latency than Gemma2 (27B) while performing respectably. As expected, the smallest model, Llama3 (8B), offered the fastest inference times but exhibited lower translation quality. Our research confirmed the importance of iterative prompt refinement for optimal LLM output. While effective, this process can be time-consuming, highlighting the need for well-crafted initial prompts to maximize efficiency in professional settings.

Regarding the platforms evaluated, Llamafile offered the lowest latency with a combined time of 1335 seconds across all translation tasks. Ollama scored a few seconds more with 1444 seconds, and GPT4All had the slowest response times with 2191 seconds. Regarding quality, numbers reveal a combined BLEU score (text translation/sentence translation) for Llamafile of 386/574, for Ollama of 364/653, and for GPT4All of 380/647. However, this is likely due to the differences in the configuration of the individual models within the various platforms than to the different platforms themselves.

However, GPT4All emerged as the most user-friendly, boasting the best interface and a wide range of settings, particularly for its LocalDocs RAG functionality. Ollama provided a comparable experience but presented greater installation and usability challenges. Llamafile, while easy to install, was the most limited in functionality. A significant challenge across all platforms was the effective implementation of retrieval-augmented generation (RAG) using a local corpus of domain-specific texts. Retrieving information from such corpora did not consistently yield satisfactory results, indicating a need for further development in this area.

In conclusion, while local LLMs are not yet ready for widespread adoption in professional translation workflows using standard hardware, the rapid advancements in both model development and hardware capabilities suggest that they hold significant potential for the future. Further research should focus on optimizing RAG techniques for local corpora, exploring the impact of specialized hardware on LLM performance within translation workflows, and developing more seamless integrations with existing translation environment tools. In addition, the development of AI agents looks promising, as these systems are able to gather information and knowledge from different sources, local as well as online. As these areas progress, local LLMs may offer translators greater control over their data, enhanced privacy, and potentially, improved efficiency and quality.

## Weblinks

Ollama: https://ollama.com; https://github.com/ollama/ollama/blob/main/README.md#quickstart
PageAssist: https://github.com/n4ze3m/page-assist
Llamafile: https://github.com/Mozilla-Ocho/Llamafile
GPT4ALL: https://gpt4all.io
HuggingFace: https://huggingface.co/models
PrivateGPT: https://github.com/zylon-ai/private-gpt
AnythingLLM: https://anythingllm.com
LM-Studio: https://lmstudio.ai
Cheshire Cat: https://cheshirecat.ai
LocalAI: https://localai.io/
Opus-Cat-MT: https://helsinki-nlp.github.io/OPUS-CAT/
DLTranslator and OmegaT: https://codeberg.org/miurahr/dltranslator/archive/main.zip
Mateo https://mateo.ivdnt.org/
Website of text sample: http://www.planoetzenhof.at/das-landgasthaus.html
Test results: http://petersandrini.net/test-results.zip